\documentclass[runningheads]{llncs}
\usepackage{graphicx}
\usepackage{tikz}
\usepackage{comment}
\usepackage{amsmath,amssymb} % define this before the line numbering.
\usepackage{color}
\usepackage[accsupp]{axessibility}  % Improves PDF readability for those with disabilities.

\usepackage{graphicx}
\usepackage{amsmath}
\usepackage{amssymb}
\usepackage{booktabs}
\usepackage{bm}
\usepackage{bbm}
\usepackage{url}
\usepackage[misc]{ifsym}
\newcommand{\boldstart}[1]{\noindent\textbf{#1}}

\renewcommand{\Omega}{\varOmega}

\begin{document}
% \linenumbers
\pagestyle{headings}
\mainmatter
\def\ECCVSubNumber{634}  % Insert your submission number here

% \title{Implicit Object Representation from Segmentation and RGB Images} % Replace with your title
\title{Object-Compositional Neural Implicit Surfaces}

\begin{comment}
\titlerunning{ECCV-22 submission ID \ECCVSubNumber} 
\authorrunning{ECCV-22 submission ID \ECCVSubNumber} 
\author{Anonymous ECCV submission}
\institute{Paper ID \ECCVSubNumber}
\end{comment}
%******************

% \begin{comment}
\titlerunning{Object-Compositional Neural Implicit Surfaces}
% \author{Qianyi Wu\inst{1}\orcidlink{0000-0001-8764-6178} \and
% Xian Liu\inst{2}\orcidlink{0000-0001-9817-7418} \and 
% Yuedong Chen\inst{1}\orcidlink{0000-0003-0943-1512} \and
% Kejie Li\inst{3}\orcidlink{0000−0001−8821−7762} \and \\
% Chuanxia Zheng\inst{1}\orcidlink{0000-0002-3584-9640} \and
% Jianfei Cai\inst{1}\orcidlink{0000-0002-9444-3763} \and 
% Jianmin Zheng\inst{4}\orcidlink{0000-0002-5062-6226}
% }

% \author{Qianyi Wu\inst{1}\orcidlink{0000-0001-8764-6178}~\textsuperscript{\Letter} \and
% Xian Liu\inst{2}\orcidlink{0000-0001-9817-7418} \and 
% Yuedong Chen\inst{1}\orcidlink{0000-0003-0943-1512} \and
% Kejie Li\inst{3}\orcidlink{0000-0001-8821-7762} \and \\
% Chuanxia Zheng\inst{1}\orcidlink{0000-0002-3584-9640} \and
% Jianfei Cai\inst{1}\orcidlink{0000-0002-9444-3763} \and 
% Jianmin Zheng\inst{4}\orcidlink{0000-0002-5062-6226}
% }
\author{Qianyi Wu\inst{1} \and
Xian Liu\inst{2}\and 
Yuedong Chen\inst{1} \and
Kejie Li\inst{3}\and \\
Chuanxia Zheng\inst{1} \and
Jianfei Cai\inst{1} \and 
Jianmin Zheng\inst{4}
}
\authorrunning{Q. Wu et al.}
\institute{Monash University 
\and 
The Chinese University of Hong Kong 
\and
University of Oxford 
\and
Nanyang Technological University\\
\email{qianyi.wu@monash.edu}
}
% \end{comment}
%******************
\maketitle

\begin{abstract}
The neural implicit representation has shown its effectiveness in novel view synthesis and high-quality 3D  reconstruction from multi-view images. However, most approaches focus on holistic scene representation yet ignore individual objects inside it, thus limiting potential downstream applications. In order to learn object-compositional representation, a few works incorporate the 2D semantic map as a cue in training to grasp the difference between objects. But they neglect the strong connections between object geometry and instance semantic information, which leads to inaccurate modeling of individual instance. This paper proposes a novel framework, \emph{ObjectSDF}, to build an object-compositional neural implicit representation with high fidelity in 3D reconstruction and object representation.
Observing the ambiguity of conventional volume rendering pipelines, we model the scene by combining the Signed Distance Functions (SDF) of individual object to exert explicit surface constraint. The key in distinguishing different instances is to revisit the strong association between an individual object's SDF and semantic label. Particularly, we convert the semantic information to a function of object SDF and develop a unified and compact representation for scene and objects. Experimental results show the superiority of \emph{ObjectSDF} framework in representing both the holistic object-compositional scene and the individual instances. Code can be found in \url{https://qianyiwu.github.io/objectsdf/}
\end{abstract}

\section{Introduction}

This paper studies the problem of efficiently learning an object-compositional 3D scene representation from posed images and semantic masks, which defines the geometry and appearance of the whole scene and individual objects as well. Such a representation characterizes the compositional nature of scenes and provides additional inherent information, thus benefiting 3D scene understanding~\cite{hassan2019resolving,nie2020total3dunderstanding,li2021moltr} and context-sensitive application tasks such as robotic manipulation~\cite{rosinol20203d,mccormac2017semanticfusion}, object editing, and AR/VR~\cite{yang2021objectnerf,yu2022unsupervised}. Learning this representation yet imposes new challenges beyond those arising in the conventional 3D scene reconstruction.

The emerging neural implicit representation rendering approaches provide promising results in novel view synthesis~\cite{mildenhall2020nerf} and 3D reconstruction~\cite{oechsle2021unisurf,yariv2021volume,wang2021neus,guo2022neural}. A typical neural implicit representation encodes scene properties into a deep network, which is trained by minimizing the discrepancies between the rendered and real RGB images from different viewpoints. For example, NeRF~\cite{mildenhall2020nerf} represents the volumetric radiance field of a scene with a neural network trained from images. The volume rendering method is used to compute pixel color, which samples points along each ray and performs $\alpha-$composition over the radiance of the sampled points. 
Despite not having direct supervision on the geometry, it is shown that neural implicit representations often implicitly learn the 3D geometry to render photorealistic images during training~\cite{mildenhall2020nerf}. However, the scene-based neural rendering in these works is mostly \emph{agnostic to individual object identities}. 

To enable the model's object-level awareness, several works are developed to encode objects' semantics into the neural implicit representation. Zhi \emph{et al.} propose an in-place scene labeling scheme~\cite{Zhi:etal:ICCV2021}, which trains the network to render not only RGB images but also 2D semantic maps. Decomposing a scene into objects can then be achieved by painting the scene-level geometric reconstruction using the predicted semantic labels. 
This workflow is not object-based modeling since the process of learning geometry is unaware of semantics. Therefore, the geometry and semantics are not strongly associated, which results in inaccurate object representation when the prediction of either geometry or semantics is bad. Yang \emph{et al.} present an object-compositional NeRF~\cite{yang2021objectnerf}, which is a unified rendering model for the scene but respecting individual object placement in the scene. The network consists of two branches: The scene branch encodes the scene geometry and appearance, and the object branch encodes each standalone object by conditioning the output only for a specific object with everything else removed. However, as proved in recent works~\cite{zhang2020nerf++,wang2021neus}, object supervision suffers from 3D space ambiguity in a clustered scene. 
It thus requires aids from extra components such as scene guidance and 3D guard masks, which are used to  distill the scene information and protect the occluded object regions.
%% typically providing the transmittance from the scene branch to guide the biased sampling of the object branch and  subtracting the visible instance space % to protect the occluded part.

Inspired by these works, we suggest modeling the object-level geometry directly to learn the geometry and semantics simultaneously so that the representation captures ``what'' and ``where'' things are in the scene. The inherent challenge is how to get the supervision for the object-level geometry from RGB images and 2D instance semantic. Unlike the semantic label for a 3D position that is well constrained by multiple 2D semantic maps using multi-view consistency, finding a direct connection between object-level geometry and the 2D semantic labels is non-trivial. In this paper, we propose a novel method called {\em ObjectSDF} for object-compositional scene representations, aiming at more accurate geometry learning in highly composite scenes and more effective extraction of individual objects to facilitate 3D scene manipulation. First, ObjectSDF represents the scene at the level of objects using a multi-layer perceptron (MLP) that outputs the Signed Distance Function (SDF) of each object at any 3D position. Note that NeRF learns a volume density field, which has difficulty in extracting a high-quality surface~\cite{zhang2021nerfactor,yariv2021volume,zhang2020nerf++,wang2021neus,zhang2022iron}. In contrast, the SDF can more accurately define surfaces and the composition of all object SDFs via the minimum operation that gives the SDF of the scene. Moreover, a density distribution can be induced by the scene SDF, which allows us to apply the volume rendering to learn an object-compositional neural implicit representation with robust network training. Second, ObjectSDF builds an explicit connection between the desired semantic field and the level set prediction, which braces the insight that the geometry of each object is strongly associated with semantic guidance. Specifically, we define the semantic distribution in 3D space as a function of each object’s SDF, which allows effective semantic guidance in learning the geometry of objects. As a result, ObjectSDF provides a unified, compact, and simple framework that can supervise the training by the input RGB and instance segmentation guidance naturally, and learn the neural implicit representation of the scene as a composition of object SDFs effectively. This is further demonstrated in our experiments.  

In summary, the paper has the following contributions: \textbf{1)} We propose a novel neural implicit surface representation using the signed distance functions {\em in an object-compositional manner}. \textbf{2)} To grasp the strong associations between object geometry and instance segmentation, we propose a simple yet effective design to incorporate the segmentation guidance organically by updating each object's SDF. \textbf{3)} We conduct experiments that demonstrate the effectiveness of the proposed method in representing individual objects and compositional scene.

\section{Related Work}
\boldstart{Neural Implicit Representation.}
Occupancy Networks~\cite{mescheder2019occupancy} and DeepSDF~\cite{park2019deepsdf} are among those pioneers who introduced the idea of encoding objects or scenes implicitly using a neural network. Such a representation can be considered as a mapping function from a 3D position to the occupancy density or SDF of the input points, which is continuous and can achieve high spatial resolution. While these works require 3D ground-truth models, Scene Representation Networks (SRN)~\cite{sitzmann2019scene} and Neural Radiance Field (NeRF)~\cite{mildenhall2020nerf} demonstrate that both geometry and appearance can be jointly learned only from multiple RGB images using multi-view consistency. Such an implicit representation idea is further used to predict the semantic segmentation label~\cite{Zhi:etal:ICCV2021,liu2022semantic}, deformation field~\cite{park2021nerfies,pumarola2021d}, high-fidelity specular reflections~\cite{verbin2021refnerf}.

This learning-by-rendering paradigm of NeRF has attracted broad interest. They also lay a foundation for many follow-up works including ours. Instead of rendering a neural radiance field, several works~\cite{yariv2020multiview,oechsle2021unisurf,luan2021unified,yariv2021volume,wang2021neus,deng2021gram} demonstrate that rendering neural implicit surfaces, where gradients are concentrated around surface regions, is able to produce a high-quality 3D reconstruction. Particularly, a recent work,  VolSDF~\cite{yariv2021volume}, combines neural implicit surface with volume rendering and produces high fidelity reconstructed surfaces.
Due to its superior modeling performance, our network is built upon VolSDF. The key difference is that VolSDF only has one SDF to model the entire scene while our work models the scene SDF as a composition of multiple object SDFs.

\boldstart{Object-Compositional Implicit Representation.} Decomposing a holistic NeRF into several parts or object-centric representations could benefit efficient rendering of radiance fields and other applications like content generation~\cite{rebain2021derf,reiser2021kilonerf,chen2022sem2nerf}. Several attempts are made to model the scene via a composition of object representations, which can be roughly categorized as category-specific~\cite{nguyen2020blockgan,guo2020object,ost2021neural,niemeyer2021giraffe} and scene-specific~\cite{yang2021objectnerf,semantic_srn,Zhi:etal:ICCV2021} methods.

The category-specific methods learn the object representation of a limited number of object categories using a large amount of training data in those categories. They have difficulty in generalizing to objects in other unseen categories. For example,  
Guo \emph{et al.}~\cite{guo2020object} propose a bottom-up method to learn one scattering field per object, which enables rendering scenes with moving objects and lights. 
Ost \emph{et al.}~\cite{ost2021neural} use a neural scene graph to represent dynamic scenes and particularly decompose objects in a street view dataset. 
Niemeyer and Geiger~\cite{niemeyer2021giraffe} propose GIRAFFE that conditions latent codes to get object-centric NeRFs and thus represents scenes as compositional generative neural feature fields.

The scene-specific methods directly learn a unified neural implicit representation for the whole scene, which also respects the object placement as in the scene~\cite{Zhi:etal:ICCV2021,yang2021objectnerf}. 
Particularly, SemanticNeRF~\cite{Zhi:etal:ICCV2021} augments NeRF to estimate the semantic label for any given 3D position. A semantic head is added into the network, which is trained by comparing the rendered and real semantic maps. Although SemanticNeRF is able to predict semantic labels, it does not explicitly model each semantic entity's geometry. 
The work closest to ours is ObjectNeRF~\cite{yang2021objectnerf}, which uses a two-pathway architecture to capture the scene and object neural radiance fields. However, the design of ObjectNeRF requires a series of additional voxel feature embedding, object activation encoding, and separate modeling of the scene and object neural radiance fields to deal with occlusion issues and improve the rendering quality. In contrast, our approach is a simple and intuitive framework that uses SDF-based neural implicit surface representation and models scene and object geometry in one unified branch.

\section{Method}
\label{sec:3}
Given a set of $N$ posed images $\mathcal{A}=\{x_1,x_2,\cdots,x_N\}$ and the corresponding instance semantic segmentation masks $\mathcal{S}=\{s_1,s_2,\cdots,s_N\}$, our goal is to learn an \emph{object-compositional implicit 3D representation} %3D model 
that captures the 3D shapes and appearances of not only the whole scene $\Omega$ but also individual \emph{objects} $\mathcal{O}$ within the scene.
%by recomposing all \emph{objects} $\mathcal{O}$ within it. 
Different from the conventional 3D scene modeling which typically models the scene as a whole without distinguishing individual objects within it, we consider the 3D scene as a composition of individual objects and the background. A unified simple yet effective framework is proposed for 3D scene and object modeling, which offers a better 3D modeling and understanding via inherent scene decomposition and recomposition. 

\begin{figure}[t!]
    \centering
    \includegraphics[width=\linewidth]{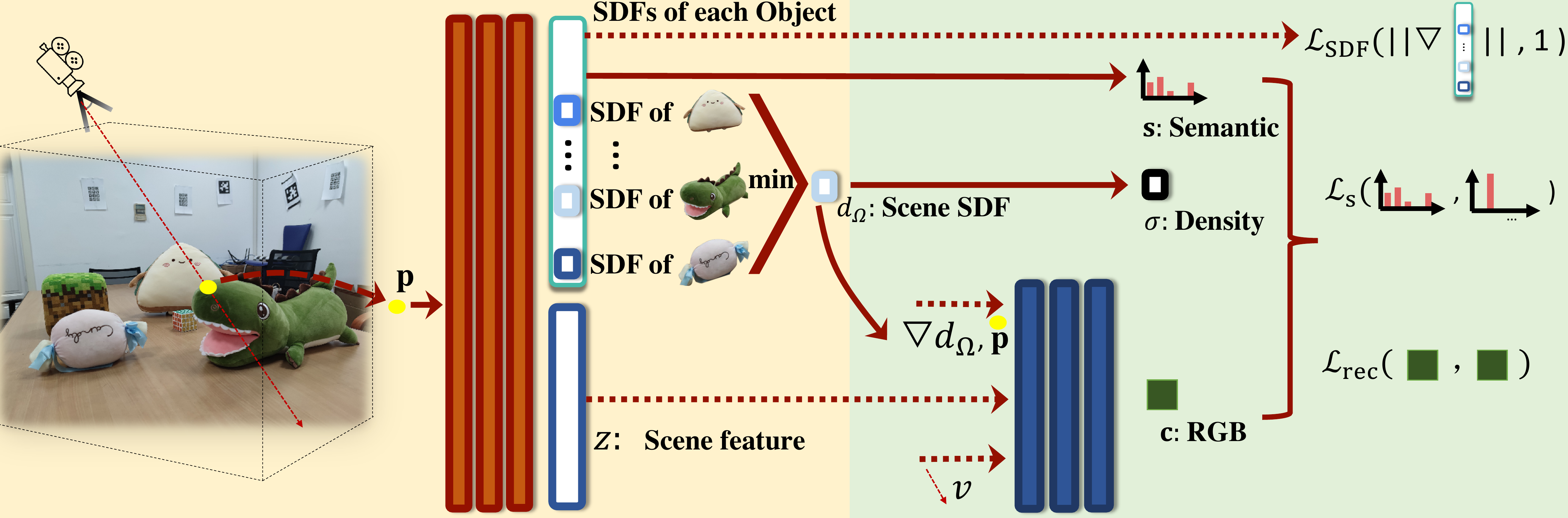}
    \caption{\textbf{Overview of our proposed \emph{ObjectSDF} framework}, consisting of two parts: an object-SDF part (left, \textcolor[rgb]{0.85, 0.80901960784314, 0.65}{yellow region}) and a scene-SDF part (right, \textcolor[rgb]{0.32941176470588, 0.6, 0.2078431372549}{green region}). The former predicts the SDF of each object, while the latter composites all object SDFs to predict the scene-level geometry and appearance.
    }
    \label{fig:overview}
\end{figure}

Fig.~\ref{fig:overview} shows the proposed \emph{ObjectSDF} framework of learning \emph{object compositional neural implicit surfaces}.
%In order to learn such a 3D scene \emph{decomposition} and \emph{recomposition}, our \emph{object-compositional} framework, illustrated in Fig~\ref{fig:overview},
It consists of an {object}-SDF part that is responsible for modeling all instances including background (Fig.~\ref{fig:overview}, yellow part) and a \emph{scene}-SDF part that recomposes the decomposed objects in the scene (Fig.~\ref{fig:overview}, green part). Note that here we use Signed Distance Function (SDF) based neural implicit surface representation  to model the geometry of the scene and objects, instead of using the popular Neural Radiance Fields (NeRF). %Our \emph{key motivation insight} is that: 
This is mainly because NeRF aims at high-quality view synthesis, not for accurate surface reconstruction, while the SDF-based neural surface representation is better for geometry modeling and SDF is also easier for the 3D composition of objects.

In the following, % describe how this is done, looking 
we first give the background of volume rendering and its combination with SDF-based neural implicit surface representation in Section~\ref{sec:3.1}. Then, we describe how to represent a scene as a composition of multiple objects within it under a unified neural implicit surface representation in Section~\ref{sec:3.2}, and emphasize our novel idea of leveraging semantic labels to supervise the modeling of individual object SDFs in Section~\ref{sec:3.3}, followed by a summary of the overall training loss in  Section~\ref{sec:3.4}.

\subsection{Background}
% introduce volume rendering, NeRF and VolSDF 
\label{sec:3.1}
\paragraph{\textbf{Volume Rendering}} %We first briefly review the volume rendering, which 
essentially takes the information from a radiance field. Considering a ray $\mathbf{r}(v) = \mathbf{o} + v\mathbf{d}$ emanated from a camera position $\mathbf{o}$ in the direction of $\mathbf{d}$, the color of the ray can be computed as an integral of the transparency $T(v)$, the density $\sigma(v)$ and the radiance $\mathbf{c}(v)$ over samples taken along near and far bounds $v_n$ and $v_f$,
\begin{equation} 
    \label{eq:nerf}
\hat{C}(\mathbf{r}) = \int_{v_n}^{v_f}T(v)\sigma(\mathbf{r}(v))\mathbf{c}(\mathbf{r}(v))dv. 
\end{equation}
This integral is approximated using a numerical quadrature~\cite{max1995optical}. The transparency function $T(v)$ represents how much light is transmitted along a ray $\mathbf{r}(v)$ and can be computed as $T(v) = \exp(-\int_{v_n}^v\sigma(\mathbf{r}(u))du)$, where the volume density $\sigma(\mathbf{p})$ is the rate that light is occluded at a point $\mathbf{p}$. Sometimes the radiance $\mathbf{c}$ may not be the function only of a ray $r(v)$, such as in~\cite{mildenhall2020nerf,yariv2020multiview}. We refer readers to \cite{kajiya1984ray} for more details about volume rendering.
% NeRF~\cite{mildenhall2020nerf} use MLPs to predict 3D density $\sigma$ and color $\mathbf{C} = (r, g, b)$ the as a function of 5D input vectors of spatial coordinates $\mathbf{p} = (x, y, z)$ and viewing direction $\mathbf{d} = (\theta, \phi)$.

\paragraph{\textbf{SDF-based Neural Implicit Surface.}} %In contrast to volume rendering, Signed Distance Function (SDF)
SDF directly characterizes the geometry at the surface. %, providing a principled way to reconstruct the surface. 
Specifically, given a scene $\mathcal{\Omega}\subset \mathbb{R}^3$, and $\mathcal{M} = \partial\mathcal{\Omega}$ is the boundary surface. The Signed Distance Function $d_{\mathcal{\Omega}}$ is defined as the distance from point $\mathbf{p}$ to the boundary $\mathcal{M}$:
\begin{equation}
    d_{\mathcal{\Omega}}(\mathbf{p}) =  (-1)^{\mathbbm{1}_{\mathcal{\Omega}}(\mathbf{p})}\min_{\mathbf{y}\in \mathcal{M}} || \mathbf{p} - \mathbf{y}||_{2},
    \label{eq:sdf}
\end{equation}
where $\mathbbm{1}_{\mathcal{\Omega}}(\mathbf{p})$ is the indicator denoting whether $\mathbf{p}$ belongs to the scene $\mathcal{\Omega}$ or not. If the point is outside the scene, $\mathbbm{1}_{\mathcal{\Omega}}(\mathbf{p})$ returns $0$; otherwise returns $1$. Typically, the standard %Euclidean 
$l_2$-norm is used to compute the distance. 

The latest neural implicit surface works~\cite{wang2021neus,yariv2021volume} combine SDF with neural implicit function and volume rendering for better geometry modeling, by replacing the NeRF volume density output $\sigma(\mathbf{p})$ with the SDF value $d_{\mathcal{\Omega}} (\mathbf{p})$, which
%, the predicted scene SDF value 
can be directly transferred into the density.  Following \cite{yariv2021volume}, here we model the density $\sigma(\mathbf{p})$ using a specific  tractable transformation: % of a learnable SDF $d_{\mathcal{\Omega}}$:
\begin{equation}
    \sigma(\mathbf{p}) = \alpha \mathbf{\Psi}(d_{\mathcal{\Omega}} (\mathbf{p})) = \left\{
    \begin{aligned}
    & \frac{1}{2\beta} \exp{(\frac{d_{\mathcal{\Omega}}(\mathbf{p})}{\beta})}   & \text{if}~ d_{\mathcal{\Omega}}(\mathbf{p}) \leq 0 \\
    & \frac{1}{\beta} - \frac{1}{2\beta} \exp{(\frac{-d_{\mathcal{\Omega}}(\mathbf{p})}{\beta})} & \text{if}~ d_{\mathcal{\Omega}}(\mathbf{p}) > 0
    \end{aligned}
    \right.
    \label{eq:densityconvert}
\end{equation}
where $\beta$ is a learnable parameter in our implementation.

%\subsection{Scene as Compositional Object Representation}
\subsection{The scene as object composition}
\label{sec:3.2}
Unlike the existing SDF-based neural implicit surface modeling works~\cite{wang2021neus,yariv2021volume}, which either focus on a single object or treat the entire scene as one object, we consider the scene as a composition of multiple objects and aim to model their geometries and appearances jointly. Specifically, 
given a static scene $\mathcal{\Omega}$, it can be naturally represented by the spatial composition of $k$ different objects $\{\mathcal{O}_i\subset \mathbb{R}^3| i = 1, \dots, k\}$, \emph{i.e.}, $\mathcal{\Omega} = \bigcup\limits_{i=1}^{k}\mathcal{O}_i$  (including background, as an individual object). %To achieve the 3D scene decomposition and recomposition, we propose to convert the \emph{Scene}-SDF as {object}-SDFs, which represents each object independently. 
Using the SDF representation, we denote the scene geometry by \emph{scene}-SDF $d_{\mathcal{\Omega}}(\mathbf{p})$ and the object geometry as {object}-SDF $d_{\mathcal{O}_i}(\mathbf{p})$, and their relationship 
%In particular, the relationship between \emph{scene}-SDF and {object}-SDFs 
can be derived as: for any point $\mathbf{p}\in \mathbb{R}^3$,     $d_{\mathcal{\Omega}}(\mathbf{p}) = \min_{i=1\dots k}d_{\mathcal{O}_i}(\mathbf{p})$. 
% \begin{equation}
%     d_{\mathcal{\Omega}}(\mathbf{p}) = \min_{i=1\dots k}d_{\mathcal{O}_i}(\mathbf{p}).
%     \label{eq:composite_sdf}
% \end{equation}
This is fundamentally different from~\cite{wang2021neus,yariv2021volume} that directly predict the SDF of the holistic scene $\mathcal{\Omega}$, %we convert the network output 
while our neural implicit function outputs $k$ distinct SDFs corresponding to different objects (see Fig.~\ref{fig:overview}). The \emph{scene}-SDF is just a minimum of the $k$ {object}-SDFs, which can be implemented as  %and each of them represents the SDF to corresponding object. This operation is easy to be implemented because the minimum operation is 
a particular type of pooling. 

Considering that we do not have any explicit supervision for the SDF values in any 3D position, we adopt the implicit geometric regulation loss~\cite{gropp2020implicit} to regularize each object SDF $d_{\mathcal{O}_i}$ as: 
\begin{equation}
    \mathcal{L}_{SDF} = \sum_{i=1}^{k}\mathbb{E}_{d_{\mathcal{O}_i}}(|| \nabla d_{\mathcal{O}_i}(\mathbf{p})|| - 1)^2.
    \label{eq:igr}
\end{equation}
This will also constrain the scene SDF $d_{\mathcal{\Omega}}$. Once we obtain the scene SDF $d_{\mathcal{\Omega}}$, we use Eq.~(\ref{eq:densityconvert}) to obtain the density in the holistic scene.

\begin{figure}%[h][t!]
    \centering
    \includegraphics[width=\linewidth]{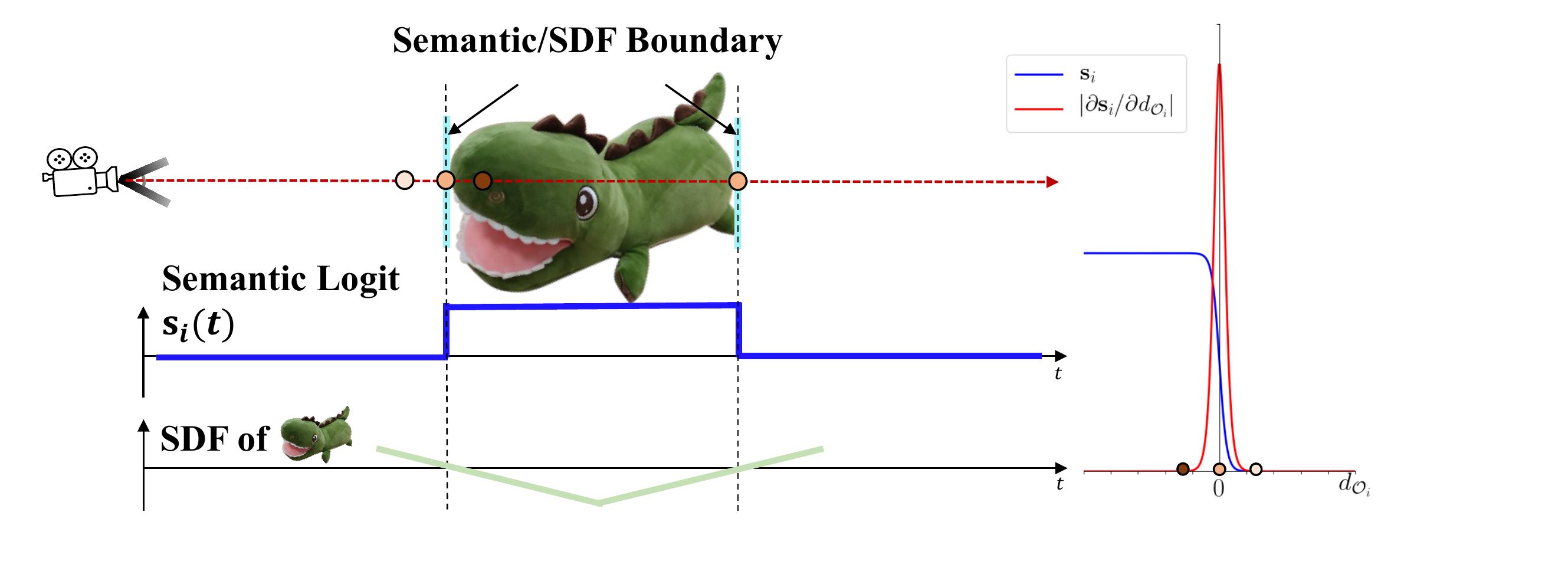}
    \caption{\textbf{Semantic as a function of {object}-SDF.} Left: the desired 3D semantic field that should satisfy the requirement, \emph{i.e.}, when the ray is crossing an object (the toy), the corresponding 3D semantic label should change rapidly. Thus, we propose to use the function \eqref{eq:si} to approximate the 3D semantic field given {object}-SDF. Right: The plot of function \eqref{eq:si} versus SDF.
    %on the right, where different curves are with different $\gamma$ values.  %The function is a smooth approximation of the semantic function.
    }
    \label{fig:semantic_field}
\end{figure}

%\subsection{Semantic Prediction as Transformed SDF}
\subsection{Leveraging semantics for learning {object}-SDFs}
\label{sec:3.3}
Although our idea of treating {scene}-SDF as a composition of multiple {object}-SDFs is simple and intuitive, 
%our object-compositional neural implicit surface representation framework is simple and intuitive, 
it is extremely challenging to learn meaningful and accurate {object}-SDFs since there is no explicit SDF supervision. The only object information we have is the given 
2D semantic masks $\mathcal{S}=\{s_1,s_2,\cdots,s_N\}$. So, the critical issue we need to address here is: \textit{How to leverage 2D instance semantic masks to guide the learning of {object}-SDFs?} %is given in our setting, we expect to incorporate it to guide the training of each object SDF.

The only existing solution we can find is the SemanticNeRF~\cite{Zhi:etal:ICCV2021}, which adds an additional head to predict a 3D ``semantic field'' $\mathbf{s}$ in the same way as predicting the radiance field $\mathbf{c}$. Then, similar to Eq.~(\ref{eq:nerf}), 2D semantic segmentation can be regarded as a volume rendering result from the 3D ``semantic field'' as:
\begin{equation}
    \hat{S}(\mathbf{r}) =  \int_{v_n}^{v_f}T(v)\sigma(\mathbf{r}(v))\mathbf{s}(\mathbf{r}(v))dv.
    \label{eq:semantic}
\end{equation}
%This operation is similar to the one used in~\cite{Zhi:etal:ICCV2021}. 
However, in our framework, $\sigma$ is transformed from \emph{scene}-SDF $d_{\mathcal{\Omega}}$, which is further obtained from {object}-SDFs. The supervision on the segmentation prediction $\hat{S}$ cannot ensure the {object}-SDFs to be meaningful.   

%Therefore, an additional semantic head as in~\cite{Zhi:etal:ICCV2021} can not force the object SDF to be meaningful. A \emph{key insight} is that semantic information is strongly associated with object geometry. 

Therefore, we turn to a new solution that represents the 3D semantic prediction $\mathbf{s}$ as a function of {object}-SDF. Our \emph{key insight} is that the semantic information is strongly associated with the object geometry. Specifically, we analyze the property of a desired 3D ``semantic field''. Considering we have $k$ objects $\{\mathcal{O}_i\subset \mathbb{R}^3| i = 1, \dots, k\}$ inside the scene including background, %and the desired semantic label $\mathbf{s}\in \mathbb{R}^k$. 
we expect that \emph{a desired 3D semantic label should maintain consistency inside one object while changing rapidly when crossing the boundary from one class to another}. Thus, we investigate the derivative of $\mathbf{s}(\mathbf{p})$ at a 3D position $\mathbf{p}$. %Because the semantic information is strongly associated with the object geometry, the semantic label of a specific class $\mathbf{s}_i$ should be a function of SDF of the specific object $d_{\mathcal{O}_i}$. 
Particularly, we inspect the norm of $\frac{\partial \mathbf{s}_i}{\partial \mathbf{p}}$:
% introduce the derivative constraint. 
\begin{equation*}
\begin{aligned}
    ||\frac{\partial \mathbf{s}_i}{\partial \mathbf{p}}|| &= ||\frac{\partial \mathbf{s}_i}{\partial d_{\mathcal{O}_i}} \cdot \frac{\partial d_{\mathcal{O}_i}}{\partial \mathbf{p}}||  &\text{(chain rule)} \\
    &\leq ||\frac{\partial \mathbf{s}_i}{\partial d_{\mathcal{O}_i}}|| \cdot || \frac{\partial d_{\mathcal{O}_i}}{\partial \mathbf{p}}||    &\text{(norm inequality)} \\
    &= ||\frac{\partial \mathbf{s}_i}{\partial d_{\mathcal{O}_i}}|| \cdot ||\nabla d_{\mathcal{O}_i}(\mathbf{p})||.
\end{aligned}
\end{equation*}
As we adopt the implicit geometric regulization loss in Eq.~(\ref{eq:igr}), $||\nabla d_{\mathcal{O}_i}(\mathbf{p})||$ should be close to $1$ after training. Therefore, the norm of $\partial \mathbf{s}_i / \partial \mathbf{p}$ should be bounded by the norm of $\partial \mathbf{s}_i/\partial d_{\mathcal{O}_i}$. In this way, we can convert the desired property of the 3D ``semantic field'' to  
%Revisiting the desire property of semantic label, we convert those constraints to
$\partial \mathbf{s}_i/\partial d_{\mathcal{O}_i}$. Considering that crossing from one class to another class means $d_{\mathcal{O}_i}$ is passing zero-level set (see Fig.~\ref{fig:semantic_field} left), we come up with a simple but effective function to satisfied the property. Concretely, we use the function: %$\mathbf{s}_i = \gamma / (1+\exp{(\gamma d_{\mathcal{O}_i})})$,
\begin{equation} \label{eq:si}
    \mathbf{s}_i = \gamma / (1+\exp{(\gamma d_{\mathcal{O}_i})}),
\end{equation}
which is a scaled sigmoid function and $\gamma$ is a hyper-parameter to control the smoothness of the function. The absolute value of $\partial \mathbf{s}_i/\partial d_{\mathcal{O}_i}$ is $\gamma^2 \exp{(\gamma d_{\mathcal{O}_i})}/(1+\exp{(\gamma d_{\mathcal{O}_i})})^2$, which meets the requirement of desired 3D semantic field, \emph{i.e.}, smooth inside the object but a rapid change at the boundary (see Fig.~\ref{fig:semantic_field} right).

This is fundamentally different from~\cite{Zhi:etal:ICCV2021}. Here we directly transform the {object}-SDF prediction $d_{\mathcal{O}_i}$ to a semantic label in 3D space. Thanks to this design, we can conduct volume rendering to convert the transformed SDF into the 2D semantic prediction using Eq.~(\ref{eq:semantic}). With the corresponding semantic segmentation mask, we minimize the cross-entropy loss $\mathcal{L}_{s}$:
\begin{equation}
    \mathcal{L}_{s} = \mathbb{E}_{\mathbf{r}\sim S}[-\log \hat{S}(\mathbf{r})].
    \label{eq:semloss}
\end{equation}
%which is chosen as a multi-class cross-entropy loss to encourage the rendered semantic labels $\hat{S}(\mathbf{r})$ to be consistent with the provided labels $S(r)$.

\subsection{Model Training}
\label{sec:3.4}
Following~\cite{mildenhall2020nerf,yariv2021volume}, we first minimize the reconstruction error between the predicted color $\hat{C}(\mathbf{r})$ and the ground-truth color $C(\mathbf{r})$ with:
\begin{equation}
    \mathcal{L}_{rec} = \mathbb{E}_{\mathbf{r}} || \hat{C}(\mathbf{r}) - C(\mathbf{r})||_1.
\end{equation}
Furthermore, we use the implicit geometric loss to regularize the SDF of each object as in Eq.~(\ref{eq:igr}). Moreover, the cross-entropy loss between the rendered semantic and ground-truth semantic is applied to guide the learning of {object}-SDFs as in Eq.~(\ref{eq:semloss}). %In summary, 
Overall, we train our model with the following three losses $\mathcal{L}_{total} = \mathcal{L}_{rec}+\lambda_1 \mathcal{L}_{s}+\lambda_2 \mathcal{L}_{SDF}$,
% \begin{equation}
%     \mathcal{L}_{total} = \mathcal{L}_{rec}+\lambda_1 \mathcal{L}_{s}+\lambda_2 \mathcal{L}_{SDF},
%     \label{eq:loss}
% \end{equation}
% We empirically set $\lambda_1$ as $0.04$ and $\lambda_2$ as $0.1$.
where $\lambda_1$ and $\lambda_2$ are two trade-off hyper-parameters. We set $\lambda_1=0.04$ and $\lambda_2=0.1$ empirically % for loss function.
\section{Experiments}
\label{sec:4}
The main purpose of our proposed method is to build an object-compositional neural implicit representation for scene rendering and object modeling. Therefore, we evaluate our approach in two real-world datasets from two aspects. Firstly, we quantitatively compare our scene representation ability with the state-of-the-art methods on standard scene rendering and modeling aspects. Then, we investigate the object representation ability of our method and compare it with NeRF-based object representation method~\cite{yang2021objectnerf}. Finally, we perform a model design ablation study to inspect the effectiveness of our framework.

\subsection{Experimental Setting}
\label{sec:4.1}
\noindent\textbf{Implementation Details.} %\footnote{Please refer to supplementary 
Our systems consists of two Multi-Layer Perceptrons (MLP). (i) The first MLP $f_{\phi}$ estimates each object SDF %for learnt geometry for each object 
as well as a scene feature $z$ of dimension $256$ for further rendering branch,  i.e., $f_{\phi}(\mathbf{p}) = [d_{\mathcal{O}_1}(\mathbf{p}), \dots, d_{\mathcal{O}_K}(\mathbf{p}), z(\mathbf{p})] \in \mathbb{R}^{K+256}$. $f_{\phi}$ consists of $6$ layers with $256$ channels. (ii) The second MLP $f_{\theta}$ is used to estimate the scene radiance field, which takes point position $\mathbf{p}$, point normal $\mathbf{n}$, view direction $\mathbf{d}$ and scene feature $z$ as inputs and outputs the RGB color $\mathbf{c}$, %in this place, 
i.e., $f_{\theta}(\mathbf{p}, \mathbf{n}, \mathbf{d}, z) = \mathbf{c}$.  $f_{\theta}$ consists of $4$ layers with $256$ channels. We use the geometric network initialization technique~\cite{Atzmon_2020_CVPR,yariv2020multiview} for both MLPs to initial the network weights to facilitate the learning of signed distance functions. We adopt the error-bounded sampling algorithm proposed by~\cite{yariv2021volume} to decide which points will be used in calculating volume rendering results. We also incorporate the positional encoding~\cite{mildenhall2020nerf} with $6$ levels for position $\mathbf{p}$ and $4$ levels for view direction $\mathbf{d}$ to help the model capture high frequency information of the geometry and radiance field. Our model can be trained in a single GTX 2080Ti GPU with a batch size of 1024 rays. 
We set $\beta=0.1$ in Eq.~\ref{eq:densityconvert} in the initial stage of training. %Our framework is implemented with PyTorch~\cite{paszke2019pytorch}.

\noindent\textbf{Datasets.} Following~\cite{yang2021objectnerf} and~\cite{Zhi:etal:ICCV2021}, we use two real datasets for comparisons. 

\noindent - \textbf{ToyDesk}~\cite{yang2021objectnerf} contains scenes of a desk by placing several toys with two different layouts and capturing images in $360^\circ$ by looking at the desk center. It also contains 2D instance segmentation for target objects as well as the camera pose for each image and a reconstructed mesh for each scene. 

\noindent - \textbf{ScanNet}  dataset~\cite{dai2017scannet}  contains RGB-D indoor scene scans as well as 3D segmentation annotations and projected 2D segmentation masks. In our experiments, we use the 2D segmentation masks provided in the ScanNet dataset for training, and the provided 3D meshes for  3D reconstruction evaluation.

\noindent\textbf{Comparison Baselines.}
We compare our method with the recent representative works in the realm of object-compositional neural implicit representation for the single static scene: \textbf{ObjectNeRF}~\cite{yang2021objectnerf} and \textbf{SemanticNeRF}~\cite{prajwal2020lip}. ObjectNeRF uses a two-path architecture to represent object-compositional neural radiance, where one branch is used for individual object modeling while the other is for scene representation. %This method is a closely related work to us. 
To broaden the ability of the network to capture accurate scene information, ObjectNeRF utilizes voxel features for both scene and object branches training, as in~\cite{liu2020neural}, which significantly increases the model complexity. {SemanticNeRF}~\cite{prajwal2020lip} is a NeRF-based framework that jointly predicts semantics and geometry in a single model for semantic labeling. The key design in this framework is an additional semantic prediction head extended from the NeRF backbone. Although this method does not directly represent objects, it can still extract an object by using semantic prediction. 

\noindent\textbf{Metric.} 
We employ the following metrics for evaluation: \textbf{1) PSNR} to evaluate the quality of rendering; \textbf{2) mIOU} to evaluate the semantic segmentation; and \textbf{3) Chamfer Distance (CD)} to measure the quality of reconstructed 3D geometry. %, and \textbf{4) \#params} measuring the number of network parameters of each method for comparing the model complexity. 
Besides these metrics, we also provide the number of neural network parameters (\textbf{\#params}) of each method for comparing the model complexity.
\begin{table}[t]
  \centering
%   \scriptsize
  \caption{\textbf{The quantitative results on scene representation.} We compare our method against recent SOTA methods~\cite{Zhi:etal:ICCV2021,yang2021objectnerf}, ablation designs and Ground Truth}
  \begin{tabular}{lcccccccc}
    \toprule
     & & \multicolumn{3}{c}{ToyDesk~\cite{yang2021objectnerf}} & \multicolumn{3}{c}{ScanNet~\cite{dai2017scannet}} &  \\
    \cmidrule(r){3-5} \cmidrule(r){6-8} 
    Methods & \#params & PSNR $\uparrow$ & mIOU $\uparrow$ & CD $\downarrow$ & PSNR $\uparrow$ & mIOU $\uparrow$ & CD $\downarrow$ \\
    Ground Truth (GT) & - & N/A & 1.00 & 0.00 & N/A & 1.00 & 0.00 \\
    \midrule
        % semantic nerf params number: 1259544(coarse: 629385 + fine: 629385)
     SemanticNeRF~\cite{Zhi:etal:ICCV2021} & $\sim$1.26M & 19.57 & 0.79 & 1.15  &  23.59 & \textbf{0.57} & 0.64 \\
     ObjectNeRF~\cite{yang2021objectnerf} & 1.78M(+19.20M) & 21.61 & 0.75 & 1.06 & 24.01 & 0.31 & 0.61\\
     \midrule
     VolSDF~\cite{yariv2021volume} &  \textbf{0.802M}& 22.00& - & 0.30 &  25.31 & - & 0.32 \\
     VolSDF w/  Semantic & $\sim$0.805M & 22.00 & \textbf{0.89} & 0.34 & \textbf{25.41} & 0.56 & 0.28 \\ 
     \midrule 
    %  Ours w/o text & 3.584 & \textbf{0.769} & 107.772 & 5.938 & 0.767 & 169.432\\
     \textbf{Ours} & \textbf{$\sim$0.804M} & \textbf{22.00} & \textbf{0.88} & \textbf{0.19} & \textbf{25.23} & 0.53 & \textbf{0.22} \\
    \bottomrule
  \end{tabular}
  \label{tbl:com_baseline}
\end{table}

\noindent\textbf{Comparison Settings.}
We follow the comparison settings introduced by ObjectNeRF~\cite{yang2021objectnerf} and SemanticNeRF~\cite{Zhi:etal:ICCV2021}. 
We use the same scene data used in~\cite{yang2021objectnerf} from ToyDesk and ScanNet for a fair comparison. 
To be consistent with SemanticNeRF~\cite{Zhi:etal:ICCV2021}, we predict the category semantic label rather than the instance semantic label for the quantitative evaluation on the ScanNet benchmark. 
Note that we are unable to train SemanticNeRF in the original resolution in the official codebase due to memory overflow. 
Therefore, we downscale the images of ScanNet and train all methods with the same data. We also noticed that the ground truth mesh may lack points in some regions, for which we apply the same crop setting for all methods to evaluate the 3D region of interest. 

It is worth noting that both our method and SemanticNeRF are able to produce the semantic label in the output. However, ObjectNeRF does not explicitly predict the semantic label in their framework. Therefore, we calculate the depth of each object which is computed from the volume density predicted from each object branch in ObjectNeRF~\cite{deng2021depth}. Then, we use the object with the nearest depth as the pixel semantic prediction of ObjectNeRF for calculating the mIOU metric. For scene rendering and the 3D reconstruction ability of ObjectNeRF, we adopt the result from the scene branch for evaluation. More details can be found in the supplementary.
\subsection{Scene-level Representation Ability}
\label{sec:4.4}
\begin{figure}[t!]
    \centering
    \includegraphics[width=\linewidth]{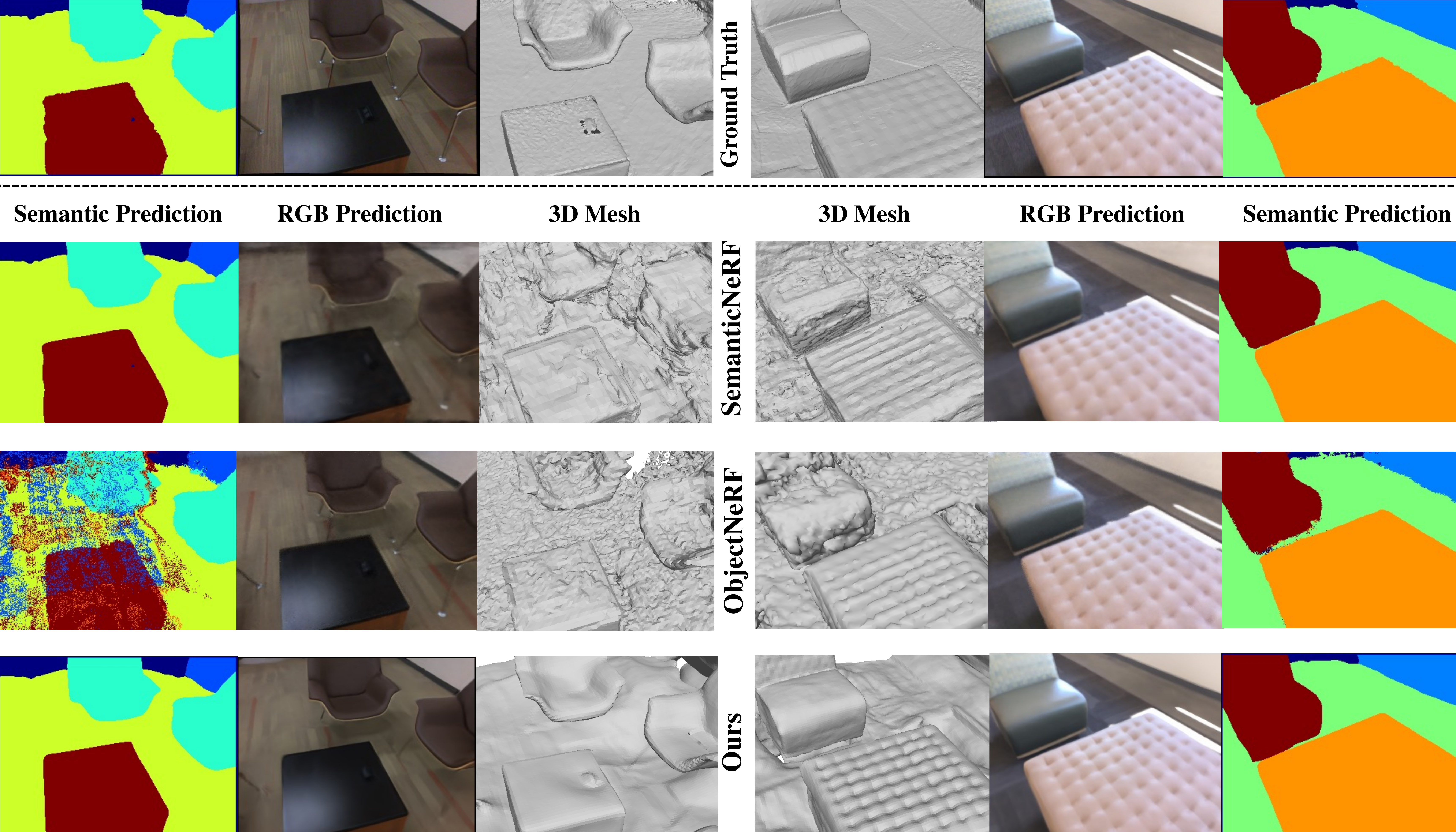}
    \caption{\textbf{Qualitative Comparison with SemanticNeRF~\cite{Zhi:etal:ICCV2021} and ObjectNeRF~\cite{yang2021objectnerf} on scene-level representation ability.} %We compare the scene representation ability of different methods. 
    We show the reconstructed meshes, predicted RGB images, and semantic masks of each method together with the ground truth results from two scenes in ScanNet.}
    \label{fig:com_sem_obj}
\end{figure}

To evaluate the scene-level representation ability, we first compare the scene rendering, object segmentation, and 3D reconstruction results. 
As shown in Tab.~\ref{tbl:com_baseline}, our framework outperforms other methods on the Toydesk benchmark and is comparable or even better than the SOTA methods on the ScanNet dataset.
The qualitative results shown in Fig.~\ref{fig:com_sem_obj} demonstrate that both our method and SemanticNeRF are able to produce fairly accurate segmentation masks. ObjectNeRF, on the other hand, renders noisy semantic masks as shown in the third row of Fig.~\ref{fig:com_sem_obj}. 
We believe the volume density predicted by the object branch is susceptible to noisy semantic prediction for points that are further from the object surface. 
Therefore, when calculating the depth of each object, it results in artifacts and leads to noisy rendering.

In terms of 3D structure reconstruction, thanks to the accurate SDF in capturing surface information, our framework can recover much more accurate geometry compared with other methods. 
We also calculate the number of model parameters of each method. 
Due to the feature volume used in ObjectNeRF, their model needs additional $19.20$M parameters. In contrast, 
the number of parameters of our model is about $0.804$M, which is about $36\%$ and $54\%$ reductions from ObjectNeRF and SemanticNeRF, respectively. 
This demonstrates the compactness and efficiency of our proposed method. 

% \noindent\textbf{Scene-Level Comparison: Show the reconstruct mesh for whole scene, predict RGB, segmentation}

\subsection{Object-level Representation Ability}
\begin{figure}[t!]
    \centering
    \includegraphics[width=\linewidth]{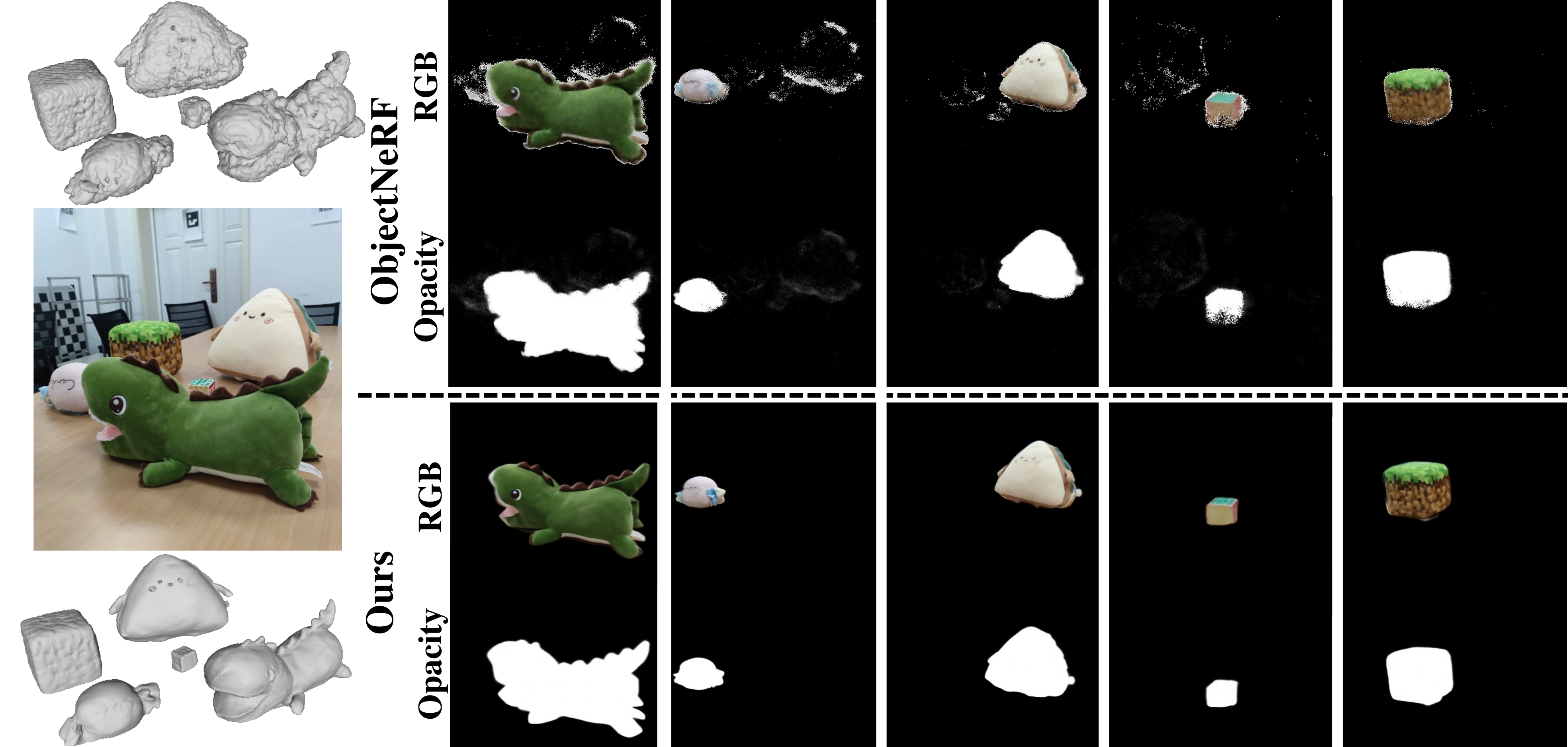}
    \caption{\textbf{Instance Results of ObjectNeRF~\cite{yang2021objectnerf} and Ours.} We show the reconstructed mesh, rendered opacity, and RGB images of different objects. %Please zoom in for better visualization. 
    % The top two rows shows the opacity of each object predicted by object branch from ObjectNeRF~\cite{yang2021objectnerf} and corresponding rendered RGB image. We also show our results on the bottom two rows. 
    % The opacity value lies in $[0, 1]$, which mapping between black to white in opacity image
    }
    \label{fig:obj_render}
\end{figure}

Besides the scene-level representation ability, our framework can naturally represent each object by selecting the specific output channel of {object}-SDFs for volume rendering. 
% Specifically, suppose we would like to represent the $i$-th object, then you can skip the minimum operation and directly use the $i$-th object's SDF as scene SDF for rendering. 
ObjectNeRF~\cite{yang2021objectnerf} can also isolate an object in a scene by computing the volume density and color of the object using the object branch network with a specific object activation code. 
We evaluate the object-level representation ability based on the quality of rendering and reconstruction of each object.
Particularly, we compare our method against ObjectNeRF on Toydesk02 which contains five toys %(excluding the background) 
in the scene as shown in Fig.~\ref{fig:obj_render}. 
We show the rendered opacity and RGB images of each toy from the same camera pose. 
It can be seen that our proposed method can render the objects more precisely with accurate opacity to describe each object.
In contrast, ObjectNeRF often renders noisy images despite utilizing the opacity loss and 3D guided mask to stop gradient during training. 
Moreover, the accurate rendering of the occluded cubes (the last two columns in Fig.~\ref{fig:obj_render}) demonstrates that our method handles occlusions much better than ObjectNeRF.
We also compare the geometry reconstructions of all the five objects on the left of Fig.~\ref{fig:obj_render}.

% \noindent\textbf{Object-Level comparison: Show the reconstruct mesh comparsion between ObjectNeRF and ObjectSDF}

\subsection{Ablation Study}
\label{sec:4.5}
% Our framework is built upon VolSDF~\cite{yariv2021volume} to develop an object-compositional neural implicit surface representation.
% Instead of modeling individual object SDFs, an alternative way to achieve the same goal is to add a semantic head to VolSDF to predict the semantic label given each 3D location, which is similar to the ``scene labeling'' approach done in SemanticNeRF~\cite{Zhi:etal:ICCV2021}.
% We name this variant as ``VolSDF w/ Semantic''.
Our framework is built upon VolSDF~\cite{yariv2021volume} to develop an object-compositional neural implicit surface representation.
Instead of modeling individual object SDFs, an alternative way to achieve the same goal is to add a semantic head to VolSDF to predict the semantic label given each 3D location, which is similar to the approach done in~\cite{Zhi:etal:ICCV2021}.
We name this variant as ``VolSDF w/ Semantic''.

\begin{figure}[t!]
    \centering
    \includegraphics[width=\linewidth]{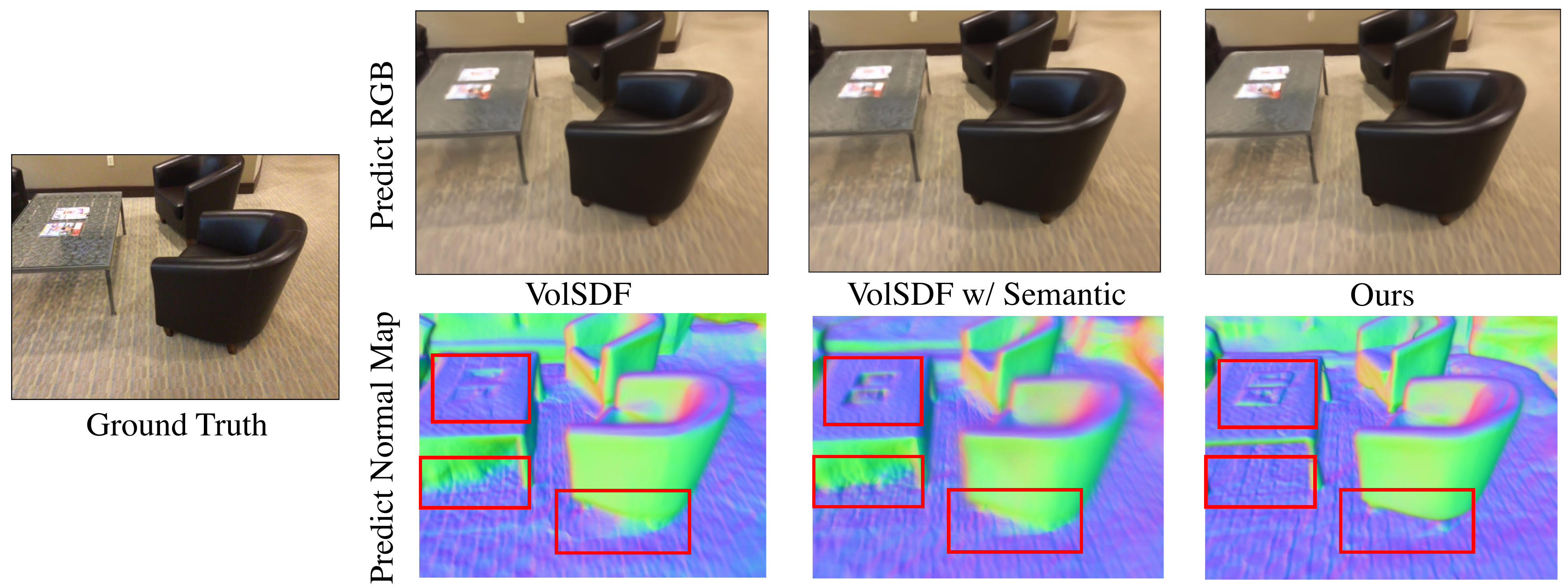}
    \caption{\textbf{Ablation study results on scene representation ability}. We show the rendered RGB image and rendered normal map together with ground truth image.
    }
    \label{fig:ablation}
\end{figure}

We first evaluate the scene-level representation ability between our method and the variant ``VolSDF w/ Semantic'' in Tab.~\ref{tbl:com_baseline}. %~\ref{tbl:ablation}.
For completeness, we also include the vanilla VolSDF, but due to the lack of semantic head, it cannot be evaluated on mIOU. 
While the comparing methods achieve similar performance on image rendering measured by PSNR, our method excels at geometric reconstruction. This is further demonstrated in Fig.~\ref{fig:ablation}, where we render the RGB image and normal map of each method. 
From the rendered normal maps, we can see that our method captures more accurate geometry compared with the two baselines. 
For example, our method can recover the geometry of the floor and the details of the sofa legs. 
The key difference between our method and the two variants is that we directly model each object SDF inside the scene. 
This indicates that our object-compositional modeling can improve the full understanding of 3D scene both semantically and geometrically.

\begin{figure}[t!]
    \centering
    \includegraphics[width=\linewidth]{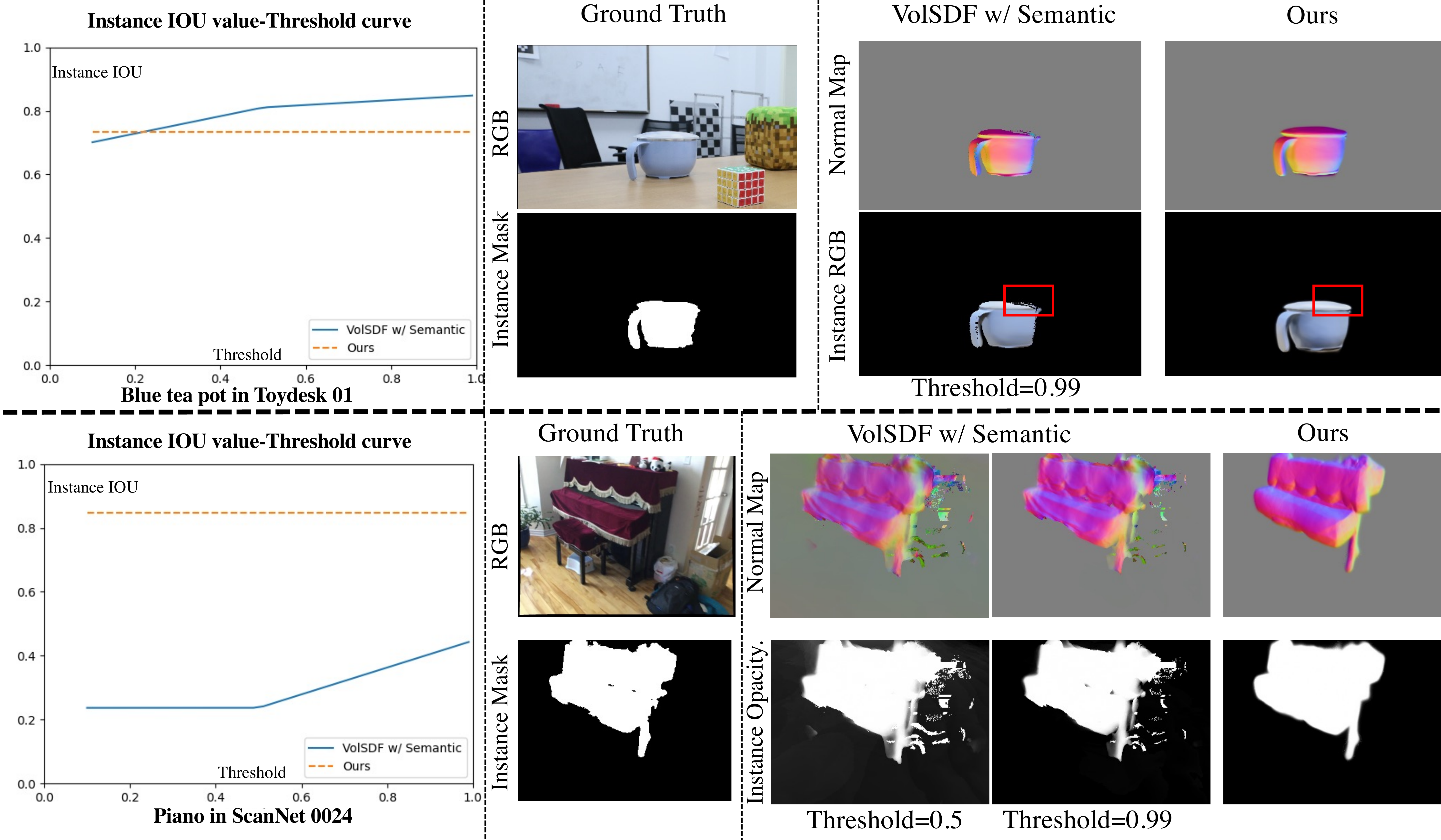}
    \caption{\textbf{Instance Results of Ours and ``VolSDF w/ Semantic''.} We show the curve between instance IOU value and semantic value threshold (left), the ground truth instance image and mask (middle), the rendered normal map, and RGB/opacity of each instance under different threshold values (right).% between ``VolSDF w/ semantic'' and ours (right)
    }
    \label{fig:ablation1}
\end{figure}

To investigate the object representation ability of ``VolSDF w/ semantic'', we %try to 
obtain an implicit object representation %given the additional semantic head. The idea is 
by using the prediction of semantic labels to determine the volume density of an object. 
In particular, given a semantic prediction in a 3D position, we can truncate the object semantic value by a threshold to decide whether to use the density to represent this object. 
We evaluate the object representation ability on two instances from ToyDesk and Scannet, respectively, in Fig.~\ref{fig:ablation1}. We choose the object, which is not occluded to extract the complete segmentation mask, and then use this mask to evaluate the semantic prediction result for each instance. 
%We adopt the IOU metric between the predicted instance mask and ground truth mask to evaluate the quality. 
Because the instance mask generated by ``VolSDF w/ Semantic'' is controlled by the semantic value threshold, we plot the curve of IOUs under different thresholds (blue line).
This reveals an inherent challenge for ``VolSDF w/ Semantic'', \emph{i.e.}, how to find a generally suitable threshold across different instances or scenes.
For instance, we notice ``VolSDF w/ Semantic'' could gain a high IOU value with a high threshold of $0.99$, but it will miss some information of the teapot (as highlighted by the red box). While using the same threshold of $0.99$ on ScanNet 0024 (bottom), it fails in separating the piano.
In contrast, our instance prediction is invariant to the threshold as shown in the yellow dash line. 
% We notice ``VolSDF w Semantic'' could gain a high IOU value with a high threshold $0.99$, but it will miss some information of the tea pot (as highlighted by red box). While using the same threshold $0.99$ on ScanNet 0024 (bottom) fail in separating the piano. 
This suggests that the separate modeling of 3D structure and semantic information is undesirable to extract accurate instance representation when either prediction is inaccurate. 
We also observe that given a fairly rough segmentation mask during training, our framework can produce a smooth and high-fidelity object representation as shown in Fig.~\ref{fig:ablation1}.

\section{Conclusion and Future Work}
We have presented an object-compositional neural implicit surface representation framework, namely \emph{ObjectSDF}, which learns the signed distance functions of all objects in a scene from the guidance of 2D instance semantic segmentation masks and RGB images using a single network. Our model unifies the object and scene representations in one framework. The main idea behind it is building a strong association between semantic information and object geometry. Extensive experimental results on two datasets have demonstrated the strong ability of our framework in both 3D scene and object representation. Future work includes applying our model for various 3D scene editing applications and efficient training of neural implicit surfaces.

\noindent \textbf{Acknowledgements} This research is partially supported by Monash FIT Start-up Grant and SenseTime Gift Fund.

\appendix
\section{Dataset}
We use the dataset as in~\cite{yang2021objectnerf} for a fair comparison. Here we give a brief introduction about these two datasets.

\noindent\textbf{ToyDesk Dataset} The ToyDesk dataset contains two image sets with 96 and 151 posed images and the corresponding instance segmentation. They capture the scene and use SfM~\cite{schonberger2016structure} and 3D reconstruction techniques~\cite{kazhdan2006poisson,xu2019multi} to recover the meshes with camera poses. And for train/test set split, they randomly sample 80$\%$ frames for training and use the rest for testing. We also use their train/testing data split as they give in the GitHub.\footnote{\url{https://github.com/zju3dv/object_nerf/tree/main/data_preparation}}.

\noindent\textbf{ScanNet Dataset} In our experiment, we choose `scene0024\_00', `scene0038\_00', `scene0113\_00' and `scene0192\_00' in ScanNet as used in ObjectNeRF~\cite{yang2021objectnerf} for fair comparison. For the experiment conducted in these data, we resize the image resolution to $320\times240$ in order to match image resolution in  SemanticNeRF~\cite{Zhi:etal:ICCV2021} and avoid the OOM issue. To match the training setting of SemanticNeRF, we use the category semantic label of ScanNet for network training and the mIOU metric evaluation of each method. 
\section{Comparison Setting Details}
We introduce the details in the comparison setting. Firstly, we will introduce the pipeline we used in calculating the semantic map of ObjectNeRF~\cite{yang2021objectnerf} since it does not explicitly produce such a result. The main principle we use in computing the semantic map of ObjectNeRF is the Z-buffer algorithm. We use the object branch in ObjectNeRF to compute the depth of each object using the following equation: $\hat{D_i}(\mathbf{r}) =  \int_{v_n}^{v_f}T_i(v)\sigma_i(\mathbf{r}(v))vdv$,
where the $T_i$, $\sigma_i$ are the object transparency and object density from $i$-th object from object branch, and $v$ is the value of depth along the ray $\mathbf{r}$. After computing the $i$-th object's depth of the ray $\hat{D_i}(\mathbf{r})$, we use the object with minimum depth value in ray $\mathbf{r}$ as the semantic label in this pixel.

We also provide the opacity computation in the experiments. The opacity is a complement probability of $T(\mathbf{r})$, which can be used to understand whether this ray be occluded in the final. The value of opacity lies in $[0, 1]$. It can be calculated as:
\begin{equation}
    \hat{O}(\mathbf{r}) =  \int_{v_n}^{v_f}T(v)\sigma(\mathbf{r}(v))dv.
\end{equation}
We adopt this value in computing the opacity map to judge the quality of rendering a single object. If the opacity of a ray is $0$, we paint it as black in the rendered image and paint it as white if it reaches $1$.
\section{Ablation study}
\begin{figure}[t!]
    \centering
    \includegraphics[width=\linewidth]{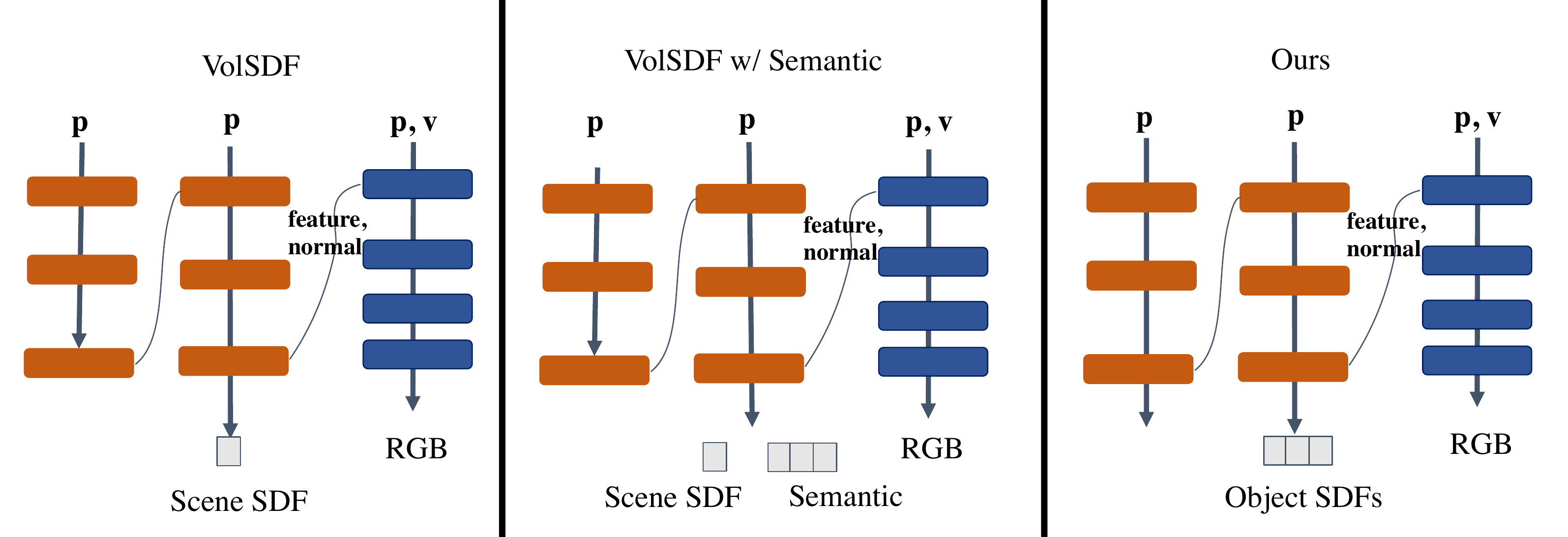}
    \vspace{-6mm}
    \caption{\textbf{Network structure of model design ablation.}, we show the network structure design in the ablation study, including ``VolSDF'', ``VolSDF w/ Semantic'', and Ours. The inputs for all methods are point position $\mathbf{p}$ and view direction $\mathbf{v}$. The difference lies in the output branch of the first MLP (orange part). 
    }
    \label{fig:supple_model}
    \vspace{-2mm}
\end{figure}
We provide more details about the ablation study related to model design. The model structure of different variants can be found in Fig.~\ref{fig:supple_model}. The difference lies in the output of the first MLP (orange part). VolSDF~\cite{yariv2021volume} predict the scene SDF and ``VolSDF w/ Semantic'' predict the scene SDF with an additional semantic prediction. However, in our framework, we directly predict the SDF of different objects and transfer them to scene SDF and semantic with a transformation function. 

We provide the details in ``VolSDF w/ Semantic" to obtain each object representation by extracting the object with a threshold. Suppose we expect to obtain the $i$-th object, we will get the semantic label $\mathbf{s}$ and volume density $\sigma$ of each point. Then we apply SoftMax operation to normalize the semantic label $\mathbf{s}$ and judge whether the $i$-th semantic label large than the given threshold $\tau$, \emph{i.e.}, $\text{SoftMax}(\mathbf{s})_i > \tau$. If the semantic label meets the requirement, we will adopt the density in this place for rendering the final result.

\begin{figure}[t!]
    \centering
    \includegraphics[width=\linewidth]{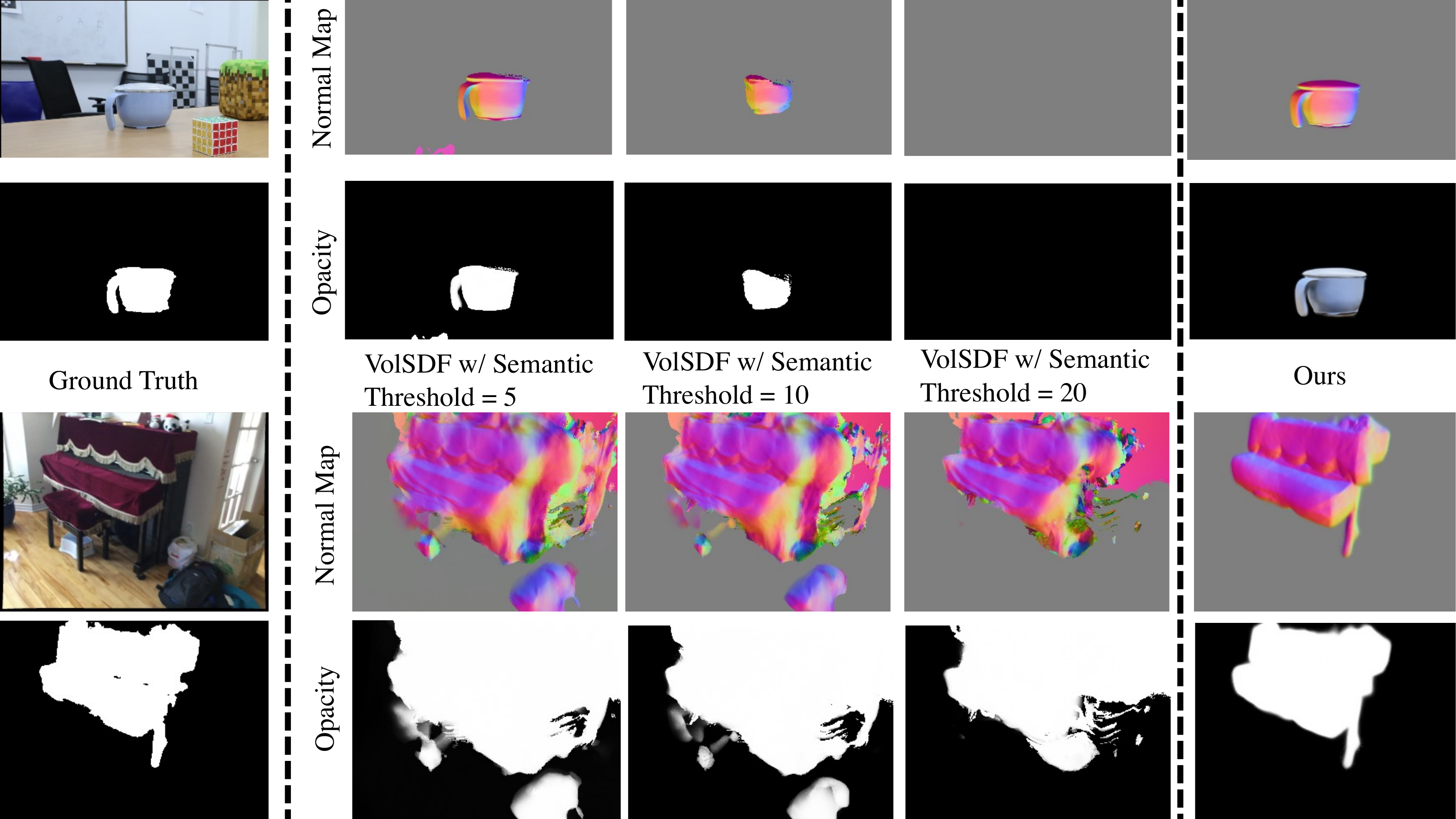}
    \vspace{-6mm}
    \caption{\textbf{Apply threshold in original semantic prediction}, we show the result that applying a threshold to extract an object from the original semantic prediction of ``VolSDF w/ Semantic". From left to right, we show the ground truth image and instance mask, the results of ``VolSDF w/ Semantic" and the results of ours
    }
    \label{fig:supple_threshold}
    \vspace{-2mm}
\end{figure}

In the supplementary, we also show more results in extracting the instance in original semantic value rather than normalized semantic value using SoftMax. The result is given in Fig.~\ref{fig:supple_threshold}. We use thresholds $5, 10$, and $20$ to extract the object. And we also notice that when the threshold is $10$, the extracted teapot is getting ruined but the piano in the bottom is far away from the ground truth. For different instances, we cannot use the same threshold to extract the object precisely for the variant ``VolSDF w/ Semantic". It again demonstrates the robustness of our proposed framework in representing objects inside the scene.

\begin{figure}[t!]
    \centering
    \includegraphics[width=\linewidth]{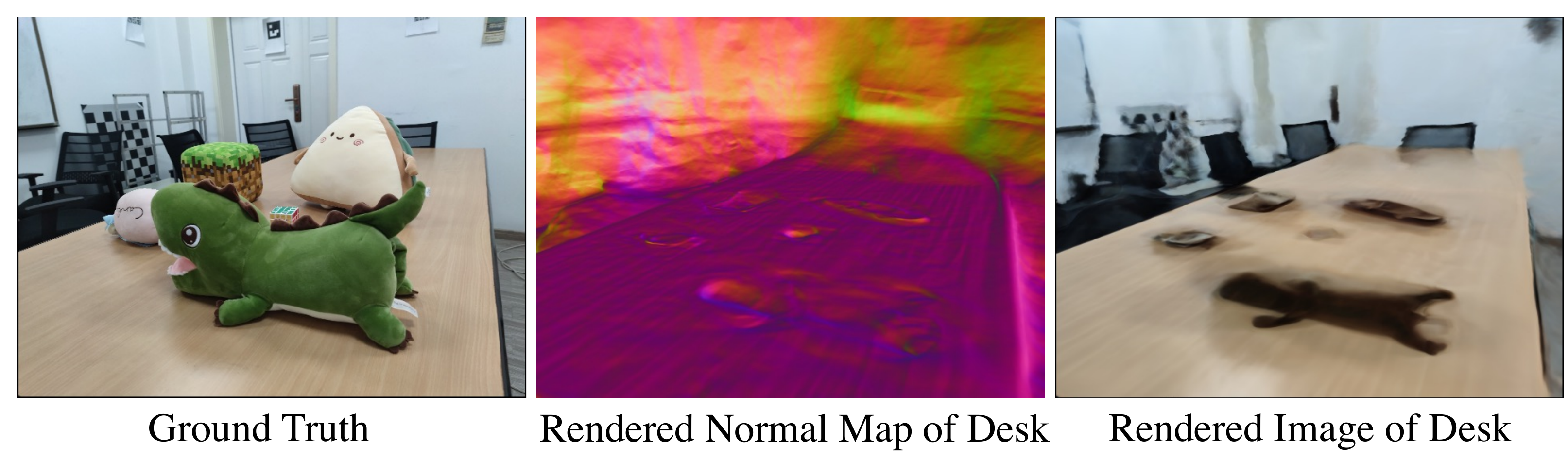}
    \vspace{-6mm}
    \caption{\textbf{Analysis of our framework}, we show the result of rendered desk result from the Toydesk dataset. From left to right, we show the ground truth image, the rendered normal map of desk (background) and the rendered image of desk. Due to the lack of observation in the bottom region of each toy, our framework cannot guarantee the reconstruction result in the invisible regions. 
    }
    \label{fig:supple_analysis}
    \vspace{-2mm}
\end{figure}

\section{Analysis of Our framework}
There still exist some limitations of our framework. As our method regarding the background as an individual object, we can also visualize the reconstructed result of the background. We give an example from ToyDesk. As shown in Fig.~\ref{fig:supple_analysis}, we notice that there are some holes in the desk region. The reason behind it is the lack of sufficient observation information in the invisible part. A possible solution to solve it is incorporating some physics constraint or causality guidance to constraint the reconstruction quality of the invisible region. We also show the rendered result of the desk, and we can observe that the texture in the invisible region also contains some artifacts. It also resulted from a lack of observation in the invisible region. Solving the reconstruction and texture issue in the invisible region is crucial for a further application like realistic scene editing. We will explore this problem in future work.

\bibliographystyle{splncs04}
\bibliography{egbib}

\clearpage


\end{document}